\documentclass[
 twocolumn,
]{ceurart}

\usepackage{listings}

\usepackage{lscape}
\begin{document}

\title{Exploring Minecraft Settlement Generators with Generative Shift Analysis}

\author[1]{Jean-Baptiste Hervé}[%
email=jbaptiste.herve@gmail.com,
]
\cormark[1]
\fnmark[1]
\address[1]{University of Hertfordshire,
  Hatfield, UK}

\author[2]{Oliver Withington}[%
orcid=0000-0002-7007-5193,
email= o.withington@qmul.ac.uk,
url=owithington.co.uk,
]
\cormark[1]
\fnmark[1]
\address[2]{Queen Mary University of London,
  London, UK}

\author{Marion Hervé}

\author[2]{Laurissa Tokarchuk}[%
email=laurissa.tokarchuk@qmul.ac.uk,
]

\author[1]{Christoph Salge}[%
email=christophsalge@gmail.com,
]

\cortext[1]{Corresponding author.}
\fntext[1]{These authors contributed equally.}

\begin{abstract}
With growing interest in Procedural Content Generation (PCG) it becomes increasingly important to develop methods and tools for evaluating and comparing alternative systems. There is a particular lack regarding the evaluation of generative pipelines, where a set of generative systems work in series to make iterative changes to an artifact. We introduce a novel method called Generative Shift for evaluating the impact of individual stages in a PCG pipeline by quantifying the impact that a generative process has when it is applied to a pre-existing artifact. We explore this technique by applying it to a very rich dataset of Minecraft game maps produced by a set of alternative settlement generators developed as part of the Generative Design in Minecraft Competition (GDMC), all of which are designed to produce appropriate settlements for a pre-existing map. While this is an early exploration of this technique we find it to be a promising lens to apply to PCG evaluation, and we are optimistic about the potential of Generative Shift to be a domain-agnostic method for evaluating generative pipelines.

\end{abstract}


\maketitle

\section{Introduction}

Procedural content generation refers to the automatic creation of pieces of content, usually called `artifacts'. Among common artifacts, we find game levels, equipment, or even 3d models. It is used to provide a lot of different assets, at any scale, without requiring to fully author.
Academic game research has studied PCG from different perspectives. Recent publications include new techniques of generation \cite{Togelius2011, summerville2018procedural}, different nature of artifacts \cite{Khalifa2017, font2013card, mahlmann2013modelling, togelius2007towards, pantaleev2012search, Doran2011}, or even design questions \cite{smith2017we}.  
One area, in particular, is the automated evaluation of generated content \cite{smith2010, summerville2018expanding, cook2016danesh, marino2015empirical, compton2016so}. However, the commonly used techniques, such as Expressive Range Analysis (ERA) \cite{smith2010}, are not well suited for intermediary steps of the generation. They also focus on the entirety of an artifact, even if its sub-components and their interactions are also good indicators of the player's perception \cite{herve2023examination}. These techniques also rely on the use of user-defined metrics, usually based on the user's intuition or previous work. But most metrics don't align with the perceived quality of generated content \cite{marino2015empirical, hervecomparing, summerville2017understanding}, nor do they have any guarantees to capture meaningful variation. Furthermore, many metrics are hard to interpret by end-users, such as game designers \cite{cook2016danesh}. The previous issues, combined with the availability of many different metrics also make the selection and visualization of metrics difficult. 

\subsection{Key Contributions and Overview}
In this paper, we make three conceptual contributions to the area of PCG system evaluation, which are as follows:

\begin{enumerate}

\item Evaluation of intermediary PCG artifacts and the impact of generative steps  - Allowing designers to interrogate the impact of individual steps in a generative process.
\item The application of dimensionality reduction to artifact metrics - Freeing designers from having to narrowly define the metrics that interest them when visualizing generative spaces.
\item ERA of locations within a single generated level, rather than of levels in totality - Allowing for the analysis of 3D environments that more closely aligns with how they are experienced by players, i.e. from a specific location in the game space.

\end{enumerate}

While each of these contributions has potential utility in isolation, they can also be combined into a single technique which we call Generative Shift Analysis (GSA). This technique allows for the qualitative and quantitative evaluation of the effect that a generator has on a base artefact in aggregate, while also allowing a user to highlight the most extremely affected individual locations within the generated environment. 

After introducing this technique we show how it can be usefully applied to gain insight from real world generative systems, in our case a set of alternative Minecraft settlement generators. We argue that GSA can be used as the basis for techniques which are better suited for evaluating PCG systems and the virtual environments they produce in the form that these systems actually exist in contemporary game development.

\section{Related work}

\subsection{Expressive Range Analysis}\label{sect_era}

Within the field of PCG, a truly qualitative evaluation of artifacts is challenging \cite{lambtutorial, compton2016so}.
The range of possibilities of a given generator is quite complex to represent - hence \emph{Expressive Range} \cite{smith2010} Analysis (ERA) is commonly employed. The Expressive Range is bounded by selected metrics, which form its dimensions. All the possible artifacts (or at least large panel of them) are plotted according to the metrics, leading to an analysis of their distribution. The analysis of the Expressive Range of a generator can be useful in order to understand its behavior and how the artifacts are spread among the dimensions. Therefore, the usefulness of an Expressive Range depends mostly on the relevance of the dimensions by which it is defined. Usually, dimensions used are automatically computed metrics applied on the whole artifact. They do not necessarily need to capture something associated with quality, but there is often an underlying assumption that higher values in certain dimensions is preferred. More importantly, the metrics should capture meaningful differences, so artifacts lying in different areas of the expressive range appear different to the relevant players. Metrics of evaluation can also be used by designers to tune the generators and optimize certain aspects of the generated artifacts.

However, ERA still has limitations.
Firstly, a computed difference between two artifacts does not necessarily lead to a perceived difference for the player \cite{compton2016so}. 
The chosen metrics themselves lack of embodiment, and several experiments have already been conducted to critically examine metrics commonly used and their relevance \cite{marino2015empirical, hervecomparing, summerville2017understanding}. 
In most cases, the Expressive Range is defined as a 2D space, mostly for ease of interpretation, but ultimately prevent in depth analysis of the interactions between several metrics \cite{summerville2018expanding}. As a consequence, picking metrics pairs that impart meaning can be time consuming and adds additional complexity \cite{withington2022a}. 
The use of the metrics is also not well defined, and depending of their nature, a "good" artifact might need to maximise a metric or target a specific value depending purpose \cite{hervecomparing}. 
Another limitation is the reliance on evaluating artifacts independently and as atomic objects, while other results suggest generated artifacts are in reality experienced as composite  \cite{herve2023examination}.
Some of ERA limitations have been addressed by further study, offering to expand it, through better representations of ER and ease of metrics selection \cite{summerville2018expanding, withington2022a, withington2023right}. But these techniques have yet to be adopted.

\subsection{Dimensionality Reduction}
One of the ideas explored in this paper is using a dimensionality reduction (DR) algorithm, principal component analysis (PCA), to compress and combine a set of quantitative metrics. DR algorithms are a family of approaches for compressing high dimensional data into lower dimensional space while maintaining as much of the information present in the original data. They are commonly used for exploratory data analysis as well as as a pre-processing step in deep learning implementations. 

DR algorithms have found use in many different contexts in prior PCG research, most commonly as a visualisation tool. They can be applied directly to the encoded representations of game content, as in the work of Justesen et.al. (\cite{justesen2018}) which used PCA to produce visualisations of the distribution of their generated levels. Withington and Tokarchuk applied PCA and other DR algorithms to generated levels to explore whether this could be used as a useful alternative to ERA without the need for metric calculation \cite{withington2022a}. Chang and Smith applied a different DR algorithm, t-SNE, to images from playthroughs of game levels rather than the encoded levels themselves to visualise reachable play states as part of their Differentia tool \cite{chang2020}. They can also be used as a generative step, as in the work of Summerville and Mateas who PCA as an intermediary step in their approach for procedurally generating Zelda levels (\cite{summerville2021}).

While this is the first application of unaugmented DR algorithms to combine sets of metrics that we are aware of, it does bear significant resemblance to another work from Withington and Tokarchuk which used Convolutional Neural Networks (CNNs) to produce compressed representations of sets of metrics for evaluating game levels \cite{withington2022b}. Our approach of using PCA on its own lacks the sophisticaiton of CNNs but has the advantage of being significantly simpler to calculate and implement.

\subsection{GDMC Competition}\label{sect_gdmc}

The GDMC is a yearly competition in which teams submit a settlement generator \cite{salge2018generative} for Minecraft, which is a computer program that can add or remove blocks from a given Minecraft maps without human intervention. All the generated settlements are then sent to the jury. The jury includes experts in various fields, such as AI, Game Design, or urbanism. Each judge scores the settlements in each of the following categories: \emph{Adaptability, Functionality, Narrative, and Aesthetic}. \emph{Adaptability} is how well the settlement is suited for its location - how well it adapts to the terrain, both on a large and small scale. \emph{Functionality} is about what affordances the settlement provides, both to the Minecraft player and the simulated villagers. It covers various aspects, such as food, production, navigability, security, etc. \emph{Narrative} reflects how well the settlement \emph{itself} tells an evocative story about its own history, and who its inhabitants are. Finally, \emph{Aesthetic} is a rating of the overall look of the settlements. In the competition, the rating of each category is computed for each generator by averaging (mean) across all judge's scores.

The GDMC has been used as a test bed for PCG evaluation studies \cite{hervé2022, herve2023examination, hervecomparing}. Minecraft is an interesting test subject in that regard. It is mostly known for its open-ended nature and has been compared to LEGO on a computer. Even though the game offers a main objective, it is mostly used as a sandbox game. Many players use the block mechanic to terraform the game world, create structures such as houses, castles, or cities, and play the game according to self-imposed goals and challenges. Since the art style and setting of Minecraft are very generic, the game affords free creation of almost any kind of artifact, with only the player's imagination setting the limits. Automatic evaluation of all the components of a Minecraft’s map is both challenging and relevant in the larger context of PCG evaluation \cite{herve2023examination}.

\subsection{Isovists and Game Spaces}\label{sect_isovist}

Isovists have been developed with the intent of capturing perception of space \cite{benedikt1979take}.
Given a bounded environment, for each point $x$, its isovist $V_x$ \cite{benedikt1979take} is the set of all the points visible from $x$.
As a result, we can compute the isovist of every positions in the space. Each isovist is characterized by a range of scalar values such as their size, shape, intervisibility, etc \cite{benedikt1979take, turner1999making, turner2001isovists, kaplan1989environmental}
These properties have been shown to have some correlation of how a given environment is experienced and appreciated \cite{wiener2007isovist, weitkamp2014validation}.

Space in general can be analyzed to understand and even predict human and social behavior  \cite{hillier1976space,ortega2005space, van2017space}. Similar analysis have also be made for video games already \cite{nitsche2008video, fernandez2011spaces}, connecting the 3D space to different layer of experience, such as narrative, gameplay, social interactions, etc.

Hervé and Salge \cite{hervé2022} already applied the isovist theory to Minecraft, with results suggesting that they could be used as new way to evaluate PCG artifact. Rather than an holistic approach, isovists allow to focus on one specific spot of the map, and extract the experience for that given location. In the rest of this paper, we will be using the data coming from the Hervé and Salge study.

\section{Methodology}
In this section we explore and explain our process for generating and synthesising isovist information from Minecraft maps, as well as our methods for using this data from pre and post settlement generation to produce useful insights about the properties of the generator. While the method used here is very specific to this domain, we argue that our techniques or at least a subset of them could be useful when analysing almost any content generator within games.

\subsection{Data Gathering and Preparation}

\subsubsection{Map Data Used - GDMC Entries}

To evaluate the novel techniques and approaches that we explore and explain in this paper, we apply them to the 2022 entries to the Generative Design in Minecraft competition. This dataset is highly useful for several reasons. Firstly the map data is publicly available (Found at gendesignmc.engineering.nyu.edu/results), meaning it can be used both by us but also by other researchers looking to modify our methods with the same data. The code used to produce them is also publicly available for most generators, meaning we can in future both explore the relationship between the architecture of the generator’s code and its output using this paper’s techniques, as well as exploring the extent to which our findings about individual generators extend to their use on alternative base maps. 

One of the most important and unique aspects of this dataset though is the presence of the base maps. As discussed in Section \ref{sect_gdmc}, the way the GDMC is organised is that entry generators have to be designed to produce settlements on any pre-existing terrain, then a set of actual maps to be used as test beds is decided by the GDMC organisers and used for all entries. Firstly this means that we can directly compare the alternative generators as they all use the same base map to generate on. More excitingly however, it lets us do pre and post generation analysis on the settlements, meaning we can drill into what specifically a given generator changed about the map. As discussed earlier this more closely mirrors the way that PCG systems are implemented in the game industry i.e as a generative pipeline of processes that iterate on a single artefact rather than the one shot generation of complete artefacts.

\subsubsection{Isovist Calculation}

As discussed in Section \ref{sect_isovist}, the concept of an isovist comes from the domain of real world space analysis. The concept was brought over to the domain of game level evaluation in ‘Automated Isovist Computation for Minecraft’ \cite{hervé2022} , and it is the method from that paper that we use to produce our base metrics in this work. For a more detailed explanation of how the isovist metrics are calculated please see \ref{AppA}, or \cite{hervé2022}. Hervé and Salge introduced 13 distinct metrics which can be calculated for a given standable location in a Minecraft map. These are all inspired by the concept of isovists and are intended to capture some aspect of the player’s experience of the space. These isovist metrics can then be calculated for either all or a subset of the standable locations within a specific area of a Minecraft map, giving us a set of locations and the associated isovist metrics for them.

\subsubsection{Isovist Metric Compression}

To facilitate visual inspection and comparison between isovist sets we wanted to reduce the dimensionality of the 13 metric dataset down to two dimensions, similar to more conventional ERA except rather than each point representing a different artefact instead they represent a location within an artefact. There are several options for producing or selecting a 2D version of the data (See Section \ref{sect_era}). In this work we compress the assembled set of metrics using PCA. PCA calculates the linear combinations of underlying features which explain the maximal variance in the data, meaning we can take and visualise the two two most explanatory components and visualise them in a 2D plot while maintaining a robust amount of variance in the underlying isovist metric data. 

Before applying PCA we pre-process the metric data by centering it, removing the mean of each metric and scaling to unit variance. This is standard practice when applying PCA to ensure no underlying variable has an outsized impact due to having a large range of values.  In our results we report the actual contributions of each underlying metric to the principal components themselves to explore the extent to which all metrics are actually represented. These principal components were calculated for the assembled set of all generated maps

\subsection{Generator Insight Methods}
In this subsection we describe the methods that we use to gather insights about individual maps and map generators, as well as methods highlighting noteworthy isovists within the generated settlements. We aim to describe the technical process for arriving at these insights, as well what useful information we gather from them about the underlying generators.

\subsubsection{Principal Component Expressive Range Analysis}

The first method we explore for gaining insight into our roster of Minecraft settlement generators is to visualise the derived isovist principal components in a scatterplot, with the position for each location in the settlement or map determined by the top two PC values for the isovist metrics at the location. This is very similar to conventional ERA in the form popularised by Smith and Whitehead \cite{smith2010} except with two distinctions. Firstly, all of the data points are derived from locations within a single map rather than each point representing a complete artefact. Secondly, we are effectively visualising diversity in terms of all 13 of the isovist metrics due to the PCA preprocessing, rather than being limited to a choice of two metrics which traditional ERA is.

As in conventional ERA these plots give us an idea of the relative diversity present in alternative content generators by how widely distributed the isovist data from a map is in this space. It can also give us insight into the extent to which two generators are producing similar output by examining the similarity between distribution shapes in these heatmaps. Heatmap similarity is a strong indicator that the effect two generators are having on a base map is broadly alike, at least in terms of their isovist metric values. While we lose the benefit of conventional ERA of highlighting exactly what combinations of metric values are likely and unlikely to be produced, we argue this is offset by the gain in being able to qualitatively understand diversity in terms of multiple metrics in the form of principal components.

\subsubsection{Principal Components as Map Overlays}

The second technique we use for gaining insights into the generators is overlaying values derived from applying PCA to the actual Minecraft maps themselves. By taking an overhead view of the map and then colour coding locations within that map based on the principal component values found at that location we can further qualitatively evaluate a generated space, and focus our investigation of it. If a designer wanted a generated space which provided a consistent experience while traversing it as a player, they might look for smoothly changing values in their maps, whereas if they wanted a map with a high amount of notable locations which stood out from their surroundings, then they would hope to see many locations with radically different values to their surroundings. Most importantly however for our use case, they support useful and direct comparison between generated settlements and the base map. Using this approach we can see both the extent to which a generator changed the base map, but also where these changes were located.

In this paper we produce these overlays by first taking an overhead view of the Minecraft map from in game. We then take the values for only the first principal component as this is the one that explains the most variance in the underlying metric data, and then localise these values on the map based on the location of the isovist that produced them. These can then be colour coded on a continuous scale based on their value. This is a direct evolution of one of the approaches used by Herve and Salge in the paper that introduced the isovist metrics to Minecraft \cite{hervé2022}, except they looked at the values for individual metrics. Note that because we are viewing the map from vertically above and only taking the highest isovist for each location, we averaged values for every coordinate pair, starting from a vertical threshold corresponding to the common ground level. This means that any influence a generator is having on the map below the top layer of blocks, such as generating cave networks beneath the surface, would likely not be visible.

\subsubsection{Calculating Generative Shift}

The final and most innovative technique we explore is what we term Generative Shift Analysis (GSA), by which we mean the change in characteristics of a location before and after a stage in a generative pipeline, in our case the generation of a settlement. More specifically we are interested in highlighting both the general trend of the shift caused by a given generative step, as well as highlighting the locations which experienced the largest shift in their isovist metrics pre and post generation. This gives designers a view of locations that experienced the biggest change in their characteristics as a result of a generative step. 

To calculate this we take the top two principal component data points for locations in a generated settlement, and pair them up the data from the matching locations in the base map. This gives us a set of pairs of 2D vectors, with one representing the combined isovist metric information for that location before a settlement was generated, and the second representing the same location afterwards. We can then calculate the 2D vector that takes us between these two points, with its magnitude acting as a heuristic for how much a location was changed by the settlement being generated in terms of its isovist metrics. This distance is what we refer to as ‘generative shift’. We then rank locations by those which experienced the largest shift, allowing us to identify the top X most transformed locations by a given generator. These locations can then be manually inspected in game. 

There is a significant amount of qualitative information that we hope to gain from applying GSA to our data set. As discussed in Section \ref{sect_era}, an issue with the majority of techniques for understanding and visualising generative spaces is that they only afford a high level abstracted view. Here instead we provide a justification for examining specific locations within a generated settlement, with a view to gaining insight into the generator as a whole. By seeing the extremes of what a generative step can do to the base artefact it gives a designer a concrete idea of how transformative a generative step can be overall. Depending on a designer’s goals this could either increase their confidence that the generator was working well, for example by not having too large an influence on an already desirable base map, or it could highlight that the generator is having too extreme an impact.

\section{Experimental Details} 

In this section we describe the set up of the pilot experiments that we ran to explore our concept of generative shift in the context of Minecraft settlement generation. 

\subsection{Isovists Calculated}

Our experiment runs on the 20 entries from 1(of 2) map from the 2021 GDMC competition. The map we have focused on is the 'volcano map' used in the 2021 GDMC as well as in the following year’s competition.

We compute isovists on surfaces where a player can stand, at a height matching the camera position for the given location and a radius $d$ of 256 blocks (the default view distance in Minecraft). For each visible coordinate, we check the type of block, if it is \emph{transparent}, and if it is a location where a player can stand \cite{hervé2022}. 

The default map, without any structure, has 92560 possible surfaces, and a single isovist takes several seconds to be computed, depending on its size. We therefore proceeded to the following subsampling of our data: for each height value (Y coordinate in Minecraft), we compute 1 isovist out of 10 possibles, picked randomly. This subsampling significantly reduces the computation time, and previous test showed it wasn't impacting the global results \cite{hervé2022}.

\subsection{Highlighted Maps}

\begin{figure*}
  \caption{Visualisations of the Two Maps Selected for Examination from the 2021 GDMC. Generator 6 with the Highest Overall Aesthetic Score and Generator 15 Being the Overall Winner}
\label{fig:mapsview}
    \includegraphics[width=\textwidth]{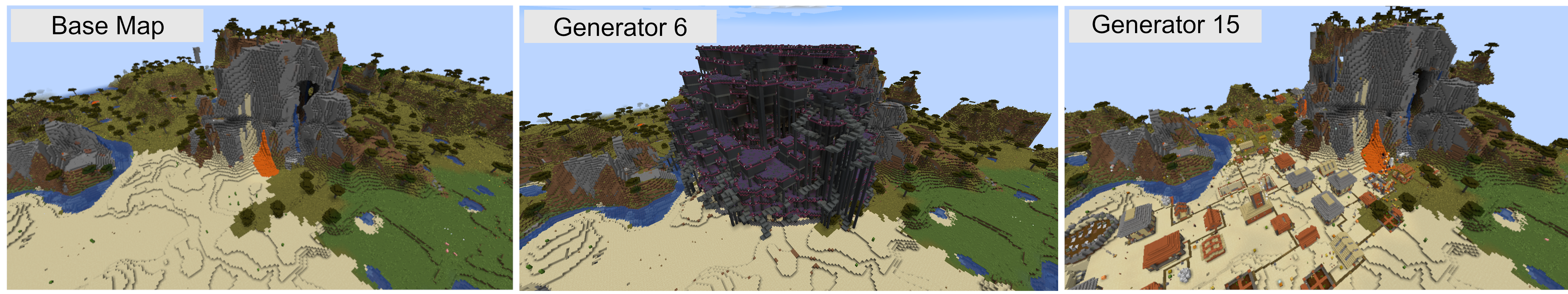}
\end{figure*}

While we have access to 20 different settlement generators, calculating the generative shift for all 20 would deliver a potentially overwhelming amount of information to dig through without further clarifying the strengths and weaknesses of the technique. As a result we opted to focus on only two generators. Generators 6 and 15 were the two highest performing generators: Generator 6 had the highest score for the Aesthetic criteria and Generator 15 was the overall highest scoring entry across all evaluation criteria. To get an idea of the character of the two generated settlements as well as the base map see Figure \ref{fig:mapsview}.

We felt that if we were only going to evaluate a subsample of generators then it made sense to select high performing generators as any insights into how they function might be more generally useful than insights gained into less high performing systems. However this decision is somewhat arbitrary, and exploring the extent to which our generative technique approach works for more diverse generators is a natural extension of this work.

\subsection{Location Selection}

To reduce the compute and processing time required to pair up and calculate the generative shift between isovists from the same location in the base map and generated settlements, we opted to only find pairings for 2\% of map locations and then calculate the generative shift for these only. This 2\% was selected stochastically during the pairing process. This sub-sampling was only required in this case due to the specific process used for gathering isovist data which was inherited from \cite{hervé2022}. This did not produce data that was easy to pair in both the base maps and generated settlements, and therefore required large numbers of euclidean distance calculations to find matched pairs. A revised data gathering process which simultaneously gathered isovist data from the same location in both maps could avoid this limitation.

While the 2\% threshold was largely arbitrary and based on the resources available for our pilot experiment, the intention for GSA is that it will still give useful insights even when only sampling a fraction of possible locations. In fact for most 3D game domains which are structured as continuous spaces rather than the large cube voxels of Minecraft, some form of sub-sampling would always be required.

\subsection{Software}

The code for extracting the isovist data from the Minecraft maps was written in Python, and relies on the GDMC's framework (https://github.com/avdstaaij/gdpc) in order to collect game's data, and Cuda for computational acceleration (https://developer.nvidia.com/cuda-python).

The code for pairing up isovists, compressing their metrics to a 2D vector using PCA, calculating their generative shift and visualising it was written in Python and is available at: (https://github.com/KrellFace/gen\_shift\_analysis). To apply PCA to the isovist metrics we used the scikit-learn Python package (https://scikit-learn.org).

\section{Results and Discussion}

\begin{figure*}
\centering
  \caption{Combined illustration of the generative shift caused by generators 6 and 15. Shift in a sub-sample of isovists displayed in top left of each section, with the 5 largest shifts highlighted in red. The locations of these isovists highlighted on overlays of PC 1 values onto the map itself. Screenshots of the locations before and after generation shown in the bottom half}
\label{fig:Gen_Shift}
    \includegraphics[width=\textwidth]{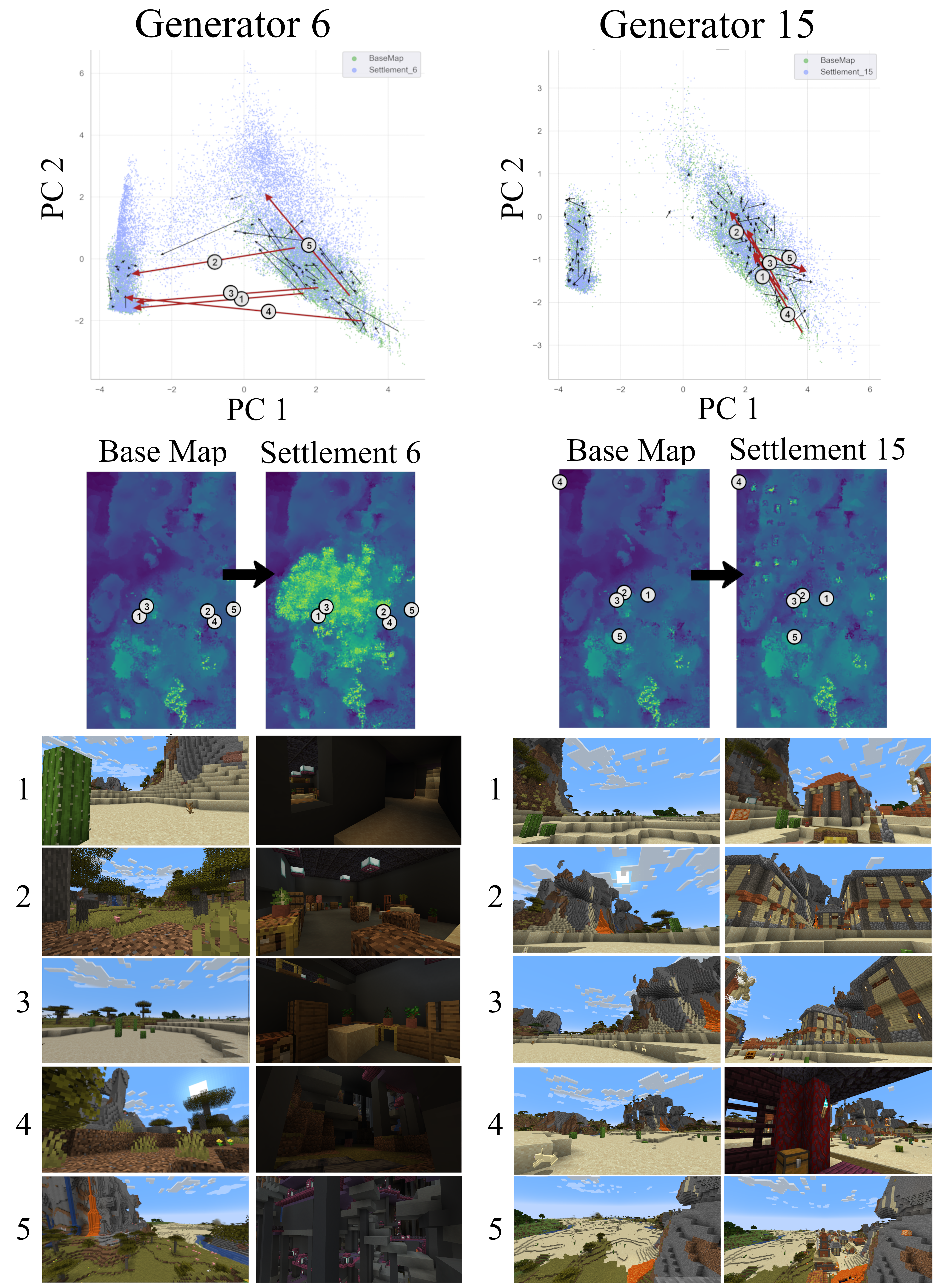}
\end{figure*}

\begin{table}[]
  \caption{Contributions of each isovist metric to the top 2 principal components, calculated for all 20 generated settlements plus base map. Presented to 3sf}
  \label{tab:pca_metrics}
\begin{tabular}{l|l|l|}
\cline{2-3}
                                                 & \textbf{PC-1} & \textbf{PC-2} \\ \hline
\multicolumn{1}{|l|}{\textbf{Area}}              & 0.291         & -0.252        \\ \hline
\multicolumn{1}{|l|}{\textbf{Perimeter}}         & 0.336         & -0.186        \\ \hline
\multicolumn{1}{|l|}{\textbf{Diversity}}         & 0.287         & -0.0610       \\ \hline
\multicolumn{1}{|l|}{\textbf{var\_Radials}}      & 0.184         & 0.387         \\ \hline
\multicolumn{1}{|l|}{\textbf{mean\_Radials}}     & 0.370         & 0.0913        \\ \hline
\multicolumn{1}{|l|}{\textbf{Roundness}}         & -0.123        & -0.114        \\ \hline
\multicolumn{1}{|l|}{\textbf{Openness}}          & 0.311         & -0.149        \\ \hline
\multicolumn{1}{|l|}{\textbf{Clutter}}           & -0.343        & 0.0256        \\ \hline
\multicolumn{1}{|l|}{\textbf{Reachability}}      & 0.0784        & 0.554         \\ \hline
\multicolumn{1}{|l|}{\textbf{Occlusivity}}       & 0.0315        & -0.558        \\ \hline
\multicolumn{1}{|l|}{\textbf{DriftLength}}       & 0.319         & 0.174         \\ \hline
\multicolumn{1}{|l|}{\textbf{VistaLength}}       & 0.359         & 0.160         \\ \hline
\multicolumn{1}{|l|}{\textbf{RealPerimeterSize}} & 0.287         & -0.170        \\ \hline
\end{tabular}
\end{table}

This section discusses the strengths, weaknesses and interesting aspects of the approach we have introduced and explored, as well as highlighting valuable directions for future inquiry that we intend to pursue ourselves or that we feel would be useful for the field.

On the application of PCA to sets of metrics, we found this to be intuitive and easy to apply, and we believe it could be useful in other contexts in which synthesising multiple metrics is desirable, not just when applying our approach of Generative Shift. The metrics themselves proved relatively compressible despite all being relevant, with the Top 2 PCs explaining over 65\% of the total variance in the data (PC 1 explained 52\% variance and PC 2 13.3\%). Being able to combine multiple phenotypic metrics let us understand the changes caused by a generator in a more holistic way and it directed us to a more diverse set of significantly altered locations (Figures \ref{fig:Gen_Shift}) than we believe would have been discoverable with a smaller set. It also had the unexpected benefit of confirming to an extent that all 13 of the isovist metrics were capturing different aspects of location diversity, as we can see in the relatively even contribution between different metrics in Table \ref{tab:pca_metrics}.

However, the significant downside of applying PCA in this way is that it makes the generative shift plots themselves and the data from aggregating them harder to interpret. Two locations could undergo shifts with very similar vectors as a result of completely different changes to their underlying metrics, and it is only when looking at the metrics or locations themselves that this may become obvious. It also has the arguable weakness of presuming that all of the metrics are equally valuable, which may well not be the case. 
The process for calculating generative shift was straightforward, though the process for finding matched locations in the base maps and generated maps was time consuming. This issue could be avoided in future implementations by gathering the metric data from both maps simultaneously and pairing up locations at that point rather than doing it post-hoc. The flow plots produced (Figure \ref{fig:Gen_Shift}) are easy to read and can quickly give an impression of the changes caused by a generative step, as well as the variance within these changes. Again, our use of PCA to compress the underlying metrics does make the plots less immediately interpretable, but the same technique could be used on pairs of unprocessed metrics instead, something we aim to investigate in future. 

In terms of the highlighted locations, we are pleased by both the significance of the changes highlighted and in the diversity present in the character of locations that were highlighted. An issue we expected might occur was that all locations highlighted would have undergone nearly identical changes, thus limiting the insights gained into the generator. However this did not occur, perhaps due to our choice to only sample 2\% of the isovists. We can only take a limited amount of credit for the diversity present in the screenshots of Figure \ref{fig:Gen_Shift} however, as the generators themselves have a significant amount of aesthetic diversity (See Figure \ref{fig:mapsview}). It is worth noting that while our method successfully capture visible changes, locations from Generator 6 are less insightful, and similar from one to another. This is likely due to the more transformative nature of the generator, and perhaps more suited metrics could lead to better results, but also to its verticality, limiting the number of shared isovist with the base map.
A limitation in terms of exhibiting these shifts, but also potentially in their calculation, is that isovists are calculated for an area around a location without factoring in a players facing, whereas a screenshot can only capture the players field of view. We aimed to capture the most significant areas of change in the facings we chose, but there is naturally some subjectivity in this choice. However the impact of this limitation is less important when actually using this approach to highlight locations, as it is trivial for a player or designer investigating the largest generative shifts to inspect the whole area.

\subsection{Future Work}\label{sect:futwork}

A direction for future study is exploring whether we can quantify the amount of generative shift  that a designer wants to see produced by a system, and whether this could be used to guide a generative process. In future work we aim to explore whether such a concept could be operationalised in a generative system by using generative shift as a pseudo fitness function to produce optimal amounts of changes to a pre-existing artefact. 

Another future direction we are excited about is one which explores the relationship between player experience and smoothness of change in isovist metric values. This builds on from the map overlay heat-maps in Figure \ref{fig:Gen_Shift} and how they allow us to easily visualise how the compressed metric values shift geographically within the map. Our hypothesis is that maps with isovist metric values that only change gradually as a player navigates the space are likely to be more relaxing or perhaps even boring than maps which present the player with more sudden changes in isovist metrics when moving. This is something we aim to explore in future, possibly with user studies using this same data set.

\section{Conclusion}

In this paper we introduced Generative Shift Analysis, a novel approach for analysing generated 3D spaces through the use of combined isovist metrics, and the changes produced in them by a generative process that uses a pre-existing virtual environment as its base. While this is only an early exploration of these concepts we still conclude that this is a potentially valuable avenue for PCG research to explore and one that we are excited to expand upon.

\section*{Acknowledgements}

We thank the reviewers for providing insightful and detailed feedback.  This work was supported in part by the EPSRC Centre for Doctoral Training in Intelligent Games \& Games Intelligence (IGGI) [EP/S022325/1].

\bibliography{isovist_exag}

\begin{thebibliography}{37}
\expandafter\ifx\csname natexlab\endcsname\relax\def\natexlab#1{#1}\fi
\providecommand{\url}[1]{\texttt{#1}}
\providecommand{\href}[2]{#2}
\providecommand{\path}[1]{#1}
\providecommand{\DOIprefix}{doi:}
\providecommand{\ArXivprefix}{arXiv:}
\providecommand{\URLprefix}{URL: }
\providecommand{\Pubmedprefix}{pmid:}
\providecommand{\doi}[1]{\href{http://dx.doi.org/#1}{\path{#1}}}
\providecommand{\Pubmed}[1]{\href{pmid:#1}{\path{#1}}}
\providecommand{\bibinfo}[2]{#2}
\ifx\xfnm\relax \def\xfnm[#1]{\unskip,\space#1}\fi
\bibitem[{{Togelius} et~al.(2011){Togelius}, {Yannakakis}, {Stanley}, and {Browne}}]{Togelius2011}
\bibinfo{author}{J.~{Togelius}}, \bibinfo{author}{G.~N. {Yannakakis}}, \bibinfo{author}{K.~O. {Stanley}}, \bibinfo{author}{C.~{Browne}},
\newblock \bibinfo{title}{Search-based procedural content generation: A taxonomy and survey},
\newblock \bibinfo{journal}{IEEE Transactions on Computational Intelligence and AI in Games} \bibinfo{volume}{3} (\bibinfo{year}{2011}) \bibinfo{pages}{172--186}. \DOIprefix\doi{10.1109/TCIAIG.2011.2148116}.
\bibitem[{Summerville et~al.(2018)Summerville, Snodgrass, Guzdial, Holmg{\aa}rd, Hoover, Isaksen, Nealen, and Togelius}]{summerville2018procedural}
\bibinfo{author}{A.~Summerville}, \bibinfo{author}{S.~Snodgrass}, \bibinfo{author}{M.~Guzdial}, \bibinfo{author}{C.~Holmg{\aa}rd}, \bibinfo{author}{A.~K. Hoover}, \bibinfo{author}{A.~Isaksen}, \bibinfo{author}{A.~Nealen}, \bibinfo{author}{J.~Togelius},
\newblock \bibinfo{title}{Procedural content generation via machine learning (pcgml)},
\newblock \bibinfo{journal}{IEEE Transactions on Games} \bibinfo{volume}{10} (\bibinfo{year}{2018}) \bibinfo{pages}{257--270}.
\bibitem[{{Khalifa} et~al.(2017){Khalifa}, {Green}, {Perez-Liebana}, and {Togelius}}]{Khalifa2017}
\bibinfo{author}{A.~{Khalifa}}, \bibinfo{author}{M.~C. {Green}}, \bibinfo{author}{D.~{Perez-Liebana}}, \bibinfo{author}{J.~{Togelius}},
\newblock \bibinfo{title}{General video game rule generation},
\newblock in: \bibinfo{booktitle}{2017 IEEE Conference on Computational Intelligence and Games (CIG)}, \bibinfo{year}{2017}, pp. \bibinfo{pages}{170--177}. \DOIprefix\doi{10.1109/CIG.2017.8080431}.
\bibitem[{Font et~al.(2013)Font, Mahlmann, Manrique, and Togelius}]{font2013card}
\bibinfo{author}{J.~M. Font}, \bibinfo{author}{T.~Mahlmann}, \bibinfo{author}{D.~Manrique}, \bibinfo{author}{J.~Togelius},
\newblock \bibinfo{title}{A card game description language},
\newblock in: \bibinfo{booktitle}{European Conference on the Applications of Evolutionary Computation}, \bibinfo{organization}{Springer}, \bibinfo{year}{2013}, pp. \bibinfo{pages}{254--263}.
\bibitem[{Mahlmann(2013)}]{mahlmann2013modelling}
\bibinfo{author}{T.~Mahlmann}, \bibinfo{title}{Modelling and generating strategy games mechanics}, \bibinfo{publisher}{IT University of Copenhagen, Center for Computer Games Research}, \bibinfo{year}{2013}.
\bibitem[{Togelius et~al.(2007)Togelius, De~Nardi, and Lucas}]{togelius2007towards}
\bibinfo{author}{J.~Togelius}, \bibinfo{author}{R.~De~Nardi}, \bibinfo{author}{S.~M. Lucas},
\newblock \bibinfo{title}{Towards automatic personalised content creation for racing games},
\newblock in: \bibinfo{booktitle}{2007 IEEE Symposium on Computational Intelligence and Games}, \bibinfo{organization}{IEEE}, \bibinfo{year}{2007}, pp. \bibinfo{pages}{252--259}.
\bibitem[{Pantaleev(2012)}]{pantaleev2012search}
\bibinfo{author}{A.~Pantaleev},
\newblock \bibinfo{title}{In search of patterns: Disrupting rpg classes through procedural content generation},
\newblock in: \bibinfo{booktitle}{Proceedings of the The third workshop on Procedural Content Generation in Games}, \bibinfo{year}{2012}, pp. \bibinfo{pages}{1--5}.
\bibitem[{Doran and Parberry(2011)}]{Doran2011}
\bibinfo{author}{J.~Doran}, \bibinfo{author}{I.~Parberry},
\newblock \bibinfo{title}{{A prototype quest generator based on a structural analysis of quests from four MMORPGs}},
\newblock \bibinfo{journal}{ACM International Conference Proceeding Series}  (\bibinfo{year}{2011}). \DOIprefix\doi{10.1145/2000919.2000920}.
\bibitem[{Smith(2017)}]{smith2017we}
\bibinfo{author}{G.~Smith},
\newblock \bibinfo{title}{What do we value in procedural content generation?},
\newblock in: \bibinfo{booktitle}{Proceedings of the 12th International Conference on the Foundations of Digital Games}, \bibinfo{year}{2017}, pp. \bibinfo{pages}{1--2}.
\bibitem[{Smith and Whitehead(2010)}]{smith2010}
\bibinfo{author}{G.~Smith}, \bibinfo{author}{J.~Whitehead},
\newblock \bibinfo{title}{Analyzing the expressive range of a level generator},
\newblock in: \bibinfo{booktitle}{Proceedings of the 2010 {Workshop} on {Procedural} {Content} {Generation} in {Games} - {PCGames} '10}, \bibinfo{publisher}{ACM Press}, \bibinfo{address}{Monterey, California}, \bibinfo{year}{2010}, pp. \bibinfo{pages}{1--7}. \URLprefix \url{http://portal.acm.org/citation.cfm?doid=1814256.1814260}. \DOIprefix\doi{10.1145/1814256.1814260}.
\bibitem[{Summerville(2018)}]{summerville2018expanding}
\bibinfo{author}{A.~Summerville},
\newblock \bibinfo{title}{Expanding expressive range: Evaluation methodologies for procedural content generation},
\newblock in: \bibinfo{booktitle}{Proceedings of the AAAI Conference on Artificial Intelligence and Interactive Digital Entertainment}, volume~\bibinfo{volume}{14}, \bibinfo{year}{2018}.
\bibitem[{Cook et~al.(2016)Cook, Gow, and Colton}]{cook2016danesh}
\bibinfo{author}{M.~Cook}, \bibinfo{author}{J.~Gow}, \bibinfo{author}{S.~Colton},
\newblock \bibinfo{title}{Danesh: Helping bridge the gap between procedural generators and their output}  (\bibinfo{year}{2016}).
\bibitem[{Mari{\~n}o et~al.(2015)Mari{\~n}o, Reis, and Lelis}]{marino2015empirical}
\bibinfo{author}{J.~Mari{\~n}o}, \bibinfo{author}{W.~Reis}, \bibinfo{author}{L.~Lelis},
\newblock \bibinfo{title}{An empirical evaluation of evaluation metrics of procedurally generated mario levels},
\newblock in: \bibinfo{booktitle}{Proceedings of the AAAI Conference on Artificial Intelligence and Interactive Digital Entertainment}, volume~\bibinfo{volume}{11}, \bibinfo{year}{2015}.
\bibitem[{Compton(2016)}]{compton2016so}
\bibinfo{author}{K.~Compton},
\newblock \bibinfo{title}{So you want to build a generator}  (\bibinfo{year}{2016}).
\bibitem[{Herv{\'e} et~al.(2023)Herv{\'e}, Salge, and Warpefelt}]{herve2023examination}
\bibinfo{author}{J.-B. Herv{\'e}}, \bibinfo{author}{C.~Salge}, \bibinfo{author}{H.~Warpefelt},
\newblock \bibinfo{title}{An examination of the hidden judging criteria in the generative design in minecraft competition}  (\bibinfo{year}{2023}).
\bibitem[{Herv{\'e} and Salge(2021)}]{hervecomparing}
\bibinfo{author}{J.-B. Herv{\'e}}, \bibinfo{author}{C.~Salge},
\newblock \bibinfo{title}{Comparing pcg metrics with human evaluation in minecraft settlement generation}  (\bibinfo{year}{2021}).
\bibitem[{Summerville et~al.(2017)Summerville, Mari{\~n}o, Snodgrass, Onta{\~n}{\'o}n, and Lelis}]{summerville2017understanding}
\bibinfo{author}{A.~Summerville}, \bibinfo{author}{J.~R. Mari{\~n}o}, \bibinfo{author}{S.~Snodgrass}, \bibinfo{author}{S.~Onta{\~n}{\'o}n}, \bibinfo{author}{L.~H. Lelis},
\newblock \bibinfo{title}{Understanding mario: an evaluation of design metrics for platformers},
\newblock in: \bibinfo{booktitle}{Proceedings of the 12th international conference on the foundations of digital games}, \bibinfo{year}{2017}, pp. \bibinfo{pages}{1--10}.
\bibitem[{Lamb et~al.(2018)Lamb, Brown, and Clarke}]{lambtutorial}
\bibinfo{author}{C.~Lamb}, \bibinfo{author}{D.~G. Brown}, \bibinfo{author}{C.~L.~A. Clarke},
\newblock \bibinfo{title}{Evaluating computational creativity: An interdisciplinary tutorial},
\newblock \bibinfo{journal}{ACM Comput. Surv.} \bibinfo{volume}{51} (\bibinfo{year}{2018}). \URLprefix \url{https://doi.org/10.1145/3167476}. \DOIprefix\doi{10.1145/3167476}.
\bibitem[{Withington and Tokarchuk(2022)}]{withington2022a}
\bibinfo{author}{O.~Withington}, \bibinfo{author}{L.~Tokarchuk},
\newblock \bibinfo{title}{Compressing and {Comparing} the {Generative} {Spaces} of {Procedural} {Content} {Generators}},
\newblock in: \bibinfo{booktitle}{2022 {IEEE} {Conference} on {Games} ({CoG})}, \bibinfo{publisher}{IEEE}, \bibinfo{address}{Beijing, China}, \bibinfo{year}{2022}, pp. \bibinfo{pages}{143--150}. \URLprefix \url{https://ieeexplore.ieee.org/document/9893615/}. \DOIprefix\doi{10.1109/CoG51982.2022.9893615}.
\bibitem[{Withington and Tokarchuk(2023)}]{withington2023right}
\bibinfo{author}{O.~Withington}, \bibinfo{author}{L.~Tokarchuk},
\newblock \bibinfo{title}{The right variety: Improving expressive range analysis with metric selection methods},
\newblock in: \bibinfo{booktitle}{Proceedings of the 18th International Conference on the Foundations of Digital Games}, \bibinfo{year}{2023}, pp. \bibinfo{pages}{1--11}.
\bibitem[{Justesen et~al.(2018)Justesen, Torrado, Bontrager, Khalifa, Togelius, and Risi}]{justesen2018}
\bibinfo{author}{N.~Justesen}, \bibinfo{author}{R.~R. Torrado}, \bibinfo{author}{P.~Bontrager}, \bibinfo{author}{A.~Khalifa}, \bibinfo{author}{J.~Togelius}, \bibinfo{author}{S.~Risi},
\newblock \bibinfo{title}{Illuminating {Generalization} in {Deep} {Reinforcement} {Learning} through {Procedural} {Level} {Generation}},
\newblock \bibinfo{journal}{arXiv:1806.10729 [cs, stat]}  (\bibinfo{year}{2018}). \URLprefix \url{http://arxiv.org/abs/1806.10729}, \bibinfo{note}{arXiv: 1806.10729}.
\bibitem[{Chang and Smith(2020)}]{chang2020}
\bibinfo{author}{K.~Chang}, \bibinfo{author}{A.~Smith},
\newblock \bibinfo{title}{Differentia: {Visualizing} {Incremental} {Game} {Design} {Changes}},
\newblock \bibinfo{journal}{Proceedings of the AAAI Conference on Artificial Intelligence and Interactive Digital Entertainment} \bibinfo{volume}{16} (\bibinfo{year}{2020}) \bibinfo{pages}{175--181}. \URLprefix \url{https://ojs.aaai.org/index.php/AIIDE/article/view/7427}. \DOIprefix\doi{10.1609/aiide.v16i1.7427}.
\bibitem[{Summerville and Mateas(2021)}]{summerville2021}
\bibinfo{author}{A.~Summerville}, \bibinfo{author}{M.~Mateas},
\newblock \bibinfo{title}{Sampling {Hyrule}: {Multi}-{Technique} {Probabilistic} {Level} {Generation} for {Action} {Role} {Playing} {Games}},
\newblock \bibinfo{journal}{Proceedings of the AAAI Conference on Artificial Intelligence and Interactive Digital Entertainment} \bibinfo{volume}{11} (\bibinfo{year}{2021}) \bibinfo{pages}{63--67}. \URLprefix \url{https://ojs.aaai.org/index.php/AIIDE/article/view/12817}. \DOIprefix\doi{10.1609/aiide.v11i3.12817}.
\bibitem[{Withington and Tokarchuk(2022)}]{withington2022b}
\bibinfo{author}{O.~Withington}, \bibinfo{author}{L.~Tokarchuk}, \bibinfo{title}{Visualising {Generative} {Spaces} {Using} {Convolutional} {Neural} {Network} {Embeddings}}, \bibinfo{year}{2022}. \URLprefix \url{http://arxiv.org/abs/2210.17464}, \bibinfo{note}{arXiv:2210.17464 [cs]}.
\bibitem[{Salge et~al.(2018)Salge, Green, Canaan, and Togelius}]{salge2018generative}
\bibinfo{author}{C.~Salge}, \bibinfo{author}{M.~C. Green}, \bibinfo{author}{R.~Canaan}, \bibinfo{author}{J.~Togelius},
\newblock \bibinfo{title}{Generative design in minecraft (gdmc) settlement generation competition},
\newblock in: \bibinfo{booktitle}{Proceedings of the 13th International Conference on the Foundations of Digital Games}, \bibinfo{year}{2018}, pp. \bibinfo{pages}{1--10}.
\bibitem[{Hervé and Salge(2022)}]{hervé2022}
\bibinfo{author}{J.-B. Hervé}, \bibinfo{author}{C.~Salge},
\newblock \bibinfo{title}{Automated {Isovist} {Computation} for {Minecraft}}  (\bibinfo{year}{2022}). \URLprefix \url{http://arxiv.org/abs/2204.03752}, \bibinfo{note}{number: arXiv:2204.03752 arXiv:2204.03752 [cs]}.
\bibitem[{Benedikt(1979)}]{benedikt1979take}
\bibinfo{author}{M.~L. Benedikt},
\newblock \bibinfo{title}{To take hold of space: isovists and isovist fields},
\newblock \bibinfo{journal}{Environment and Planning B: Planning and design} \bibinfo{volume}{6} (\bibinfo{year}{1979}) \bibinfo{pages}{47--65}.
\bibitem[{Turner and Penn(1999)}]{turner1999making}
\bibinfo{author}{A.~Turner}, \bibinfo{author}{A.~Penn},
\newblock \bibinfo{title}{Making isovists syntactic: isovist integration analysis},
\newblock in: \bibinfo{booktitle}{2nd International Symposium on Space Syntax, Brasilia}, \bibinfo{organization}{Citeseer}, \bibinfo{year}{1999}.
\bibitem[{Turner et~al.(2001)Turner, Doxa, O'sullivan, and Penn}]{turner2001isovists}
\bibinfo{author}{A.~Turner}, \bibinfo{author}{M.~Doxa}, \bibinfo{author}{D.~O'sullivan}, \bibinfo{author}{A.~Penn},
\newblock \bibinfo{title}{From isovists to visibility graphs: a methodology for the analysis of architectural space},
\newblock \bibinfo{journal}{Environment and Planning B: Planning and design} \bibinfo{volume}{28} (\bibinfo{year}{2001}) \bibinfo{pages}{103--121}.
\bibitem[{Kaplan et~al.(1989)Kaplan, Kaplan, and Brown}]{kaplan1989environmental}
\bibinfo{author}{R.~Kaplan}, \bibinfo{author}{S.~Kaplan}, \bibinfo{author}{T.~Brown},
\newblock \bibinfo{title}{Environmental preference: A comparison of four domains of predictors},
\newblock \bibinfo{journal}{Environment and behavior} \bibinfo{volume}{21} (\bibinfo{year}{1989}) \bibinfo{pages}{509--530}.
\bibitem[{Wiener et~al.(2007)Wiener, Franz, Rossmanith, Reichelt, Mallot, and B{\"u}lthoff}]{wiener2007isovist}
\bibinfo{author}{J.~M. Wiener}, \bibinfo{author}{G.~Franz}, \bibinfo{author}{N.~Rossmanith}, \bibinfo{author}{A.~Reichelt}, \bibinfo{author}{H.~A. Mallot}, \bibinfo{author}{H.~H. B{\"u}lthoff},
\newblock \bibinfo{title}{Isovist analysis captures properties of space relevant for locomotion and experience},
\newblock \bibinfo{journal}{Perception} \bibinfo{volume}{36} (\bibinfo{year}{2007}) \bibinfo{pages}{1066--1083}.
\bibitem[{Weitkamp et~al.(2014)Weitkamp, van Lammeren, and Bregt}]{weitkamp2014validation}
\bibinfo{author}{G.~Weitkamp}, \bibinfo{author}{R.~van Lammeren}, \bibinfo{author}{A.~Bregt},
\newblock \bibinfo{title}{Validation of isovist variables as predictors of perceived landscape openness},
\newblock \bibinfo{journal}{Landscape and urban planning} \bibinfo{volume}{125} (\bibinfo{year}{2014}) \bibinfo{pages}{140--145}.
\bibitem[{Hillier et~al.(1976)Hillier, Leaman, Stansall, and Bedford}]{hillier1976space}
\bibinfo{author}{B.~Hillier}, \bibinfo{author}{A.~Leaman}, \bibinfo{author}{P.~Stansall}, \bibinfo{author}{M.~Bedford},
\newblock \bibinfo{title}{Space syntax},
\newblock \bibinfo{journal}{Environment and Planning B: Planning and design} \bibinfo{volume}{3} (\bibinfo{year}{1976}) \bibinfo{pages}{147--185}.
\bibitem[{Ortega-Andeane et~al.(2005)Ortega-Andeane, Jim{\'e}nez-Rosas, Mercado-Dom{\'e}nech, and Estrada-Rodr{\'\i}guez}]{ortega2005space}
\bibinfo{author}{P.~Ortega-Andeane}, \bibinfo{author}{E.~Jim{\'e}nez-Rosas}, \bibinfo{author}{S.~Mercado-Dom{\'e}nech}, \bibinfo{author}{C.~Estrada-Rodr{\'\i}guez},
\newblock \bibinfo{title}{Space syntax as a determinant of spatial orientation perception},
\newblock \bibinfo{journal}{International Journal of Psychology} \bibinfo{volume}{40} (\bibinfo{year}{2005}) \bibinfo{pages}{11--18}.
\bibitem[{van Nes and Yamu(2017)}]{van2017space}
\bibinfo{author}{A.~van Nes}, \bibinfo{author}{C.~Yamu},
\newblock \bibinfo{title}{Space syntax: A method to measure urban space related to social, economic and cognitive factors},
\newblock in: \bibinfo{booktitle}{The virtual and the real in planning and urban design}, \bibinfo{publisher}{Routledge}, \bibinfo{year}{2017}, pp. \bibinfo{pages}{136--150}.
\bibitem[{Nitsche(2008)}]{nitsche2008video}
\bibinfo{author}{M.~Nitsche}, \bibinfo{title}{Video game spaces: image, play, and structure in 3D worlds}, \bibinfo{publisher}{MIT Press}, \bibinfo{year}{2008}.
\bibitem[{Fern{\'a}ndez-Vara(2011)}]{fernandez2011spaces}
\bibinfo{author}{C.~Fern{\'a}ndez-Vara},
\newblock \bibinfo{title}{Game spaces speak volumes: Indexical storytelling},
\newblock in: \bibinfo{booktitle}{DiGRA '11 - Proceedings of the 2011 DiGRA International Conference: Think Design Play}, \bibinfo{publisher}{DiGRA/Utrecht School of the Arts}, \bibinfo{year}{2011}.

\end{thebibliography}
\appendix

\section{Set-based definitions of isovist metrics} \label{AppA}

Given a point $x$ and its isovist $V_x$, $x$ is called the centroid of $V_x$. The lines connecting $x$ and the boundary of $V_x$ are referred to as radials $l_{x,\theta}$.
$V_x$ is also defined by its the visible area ($A_x$), and its perimeter ($P_x$). It is worthwhile to note that $P_x$ is defined by "real-surfaces" which are defined by Benedidkt as “opaque, material, visible surface, humanly perceivable as such", and therefore exclude the sky or any glass surface from the computation.

A Minecraft world can be, in large part, defined by its constituent blocks. Each 3d coordinate basically contains one block, and defining the type of block present at each coordinate defines the world map. Other objects are also made up of blocks, just arranged differently. 

In order to implement our metrics, we defined a number of sets, denoted by capital letters $X$, each containing a number of unique blocks defined by their x,y, and z coordinates and their type. 

We also use the concept of \emph{headspace}, defined as the block a player avatar's head could be for a standing avatar. Every empty block, has an empty block below it, and has a \emph{standable} block below that, is a possible headspace. The list of empty blocks contains air, but also some less common blocks such as doors, carpets, etc, that still allow the player to enter those blocks. Standable blocks are similarly defined via a list, and contain most solid blocks, but exclude things like lava or water. The two empty blocks are due to the 2 unit height of a player avatar. This should capture all blocks a player could be in $H_{\forall}$. The following sets are computed for all possible headspaces $x \in H_{\forall}$
.
\begin{eqnarray}
   H_{x,d} &:=& \{\textrm{all } blocks \in H_{\forall} \textrm{ visible from } x\}  \\
   H_{x-2} &:=& \{\textrm{blocks 2 units below every } \in H_x\}  \\
   P_{x,d} &:=&\{ \textrm{all blocks visible from } x\}  \\
   Pr_{x,d} &:=&\{ \textrm{all real surfaces blocks visible from } x\}  \\
   R_{x,n} &:=&\{ \textrm{walkable blocks from $x-2$ in $n$ steps}\}
\end{eqnarray}

Visibility between two blocks is computed by ray-casting via Bresenham algorithm, and checking if the blocks along the ray are transparent, i.e. in the transparency list. Air is the main transparency provider, but notably glass provides transparency without allowing avatars to pass. The parameter $d$ is used to limit the max length for this ray cast. We also compute the length of the many rays (radials) cast during the visibility computations to compute:

\begin{eqnarray}
\textrm{var Radials} &:=& Var(l_{x,\theta})\\
\textrm{mean Radials} &:=& Mean(l_{x,\theta})\\
\textrm{Drift} &:=& l_{x,Mean(\theta)}\\
\textrm{Vista Length} &:=& Max(l_{x,\theta})
\end{eqnarray}
The visible head spaces $H_{x,d}$ are basically all positions an avatar could be in and then see its head from its current position. $H_x-2$ are all the standable blocks supporting those headspaces. $P_{x,d}$ provides a perimeter of blocks that limit our view and contains all blocks visible from the current position. $R_{x,n}$ is a list of all walkable blocks, obtained by floodfilling from the standable block supporting the current position, within $n$ steps. We use usual Minecraft movement rules, that allow moving up by one block per lateral transverse, and dropping down to lower levels. Note how the features of avatar height, movement rules, and vision sensors could affect those basic sets. The following properties can now be computed by operating on those sets alone, without having to recompute them. 

\begin{eqnarray}
\textrm{Area} &:=& |H_{x,d}| \\
\textrm{Perimeter} &:=& |P_{x,d}|\\
\textrm{Diversity} &:=& c(P_{x,d})\\
\textrm{Real Perimeter} &:=& |(Pr_{x,d})|\\
\textrm{Roundness} &:=& Area / Perimeter\\
\textrm{Openness} &:=& Area / Real Perimeter\\
\textrm{Reachability} &:=& |R_{x,n}|\\
\textrm{Occlusivity} &:=& |R_{x,n} \cap H_{x-2}|/ Reachability\\
\textrm{Clutter} &:=& |H_{x-2} \cap P_{x,d}|/ Area
\end{eqnarray}
The function $c(.)$ counts how many different types of blocks are in a set. 

\end{document}